# SAFELOC: Overcoming Data Poisoning Attacks in Heterogeneous Federated Machine Learning for Indoor Localization

Akhil Singampalli, Danish Gufran and Sudeep Pasricha
Department of Electrical and Computer Engineering,
Colorado State University Fort Collins, CO, USA
{Akhil.Singampalli, Danish.Gufran, Sudeep}@colostate.edu

*Abstract*— Machine learning (ML) based indoor localization solutions are critical for many emerging applications, yet their efficacy is often compromised by hardware/software variations across mobile devices (i.e., device heterogeneity) and the threat of ML data poisoning attacks. Conventional methods aimed at countering these challenges show limited resilience to the uncertainties created by these phenomena. In response, in this paper, we introduce *SAFELOC*, a novel framework that not only minimizes localization errors under these challenging conditions but also ensures model compactness for efficient mobile device deployment. Our framework targets a distributed and co-operative learning environment that uses federated learning (FL) to preserve user data privacy and assumes heterogeneous mobile devices carried by users (just like in most real-world scenarios). Within this heterogeneous FL context, *SAFELOC* introduces a novel fused neural network architecture that performs data poisoning detection and localization, with a low model footprint. Additionally, a dynamic saliency map-based aggregation strategy is designed to adapt based on the severity of the detected data poisoning scenario. Experimental evaluations demonstrate that *SAFELOC* achieves improvements of up to 5.9× in mean localization error, 7.8× in worst-case localization error, and a 2.1× reduction in model inference latency compared to state-of-the-art indoor localization frameworks, across diverse building floorplans, mobile devices, and ML data poisoning attack scenarios.

## I. INTRODUCTION

Indoor localization systems are rapidly gaining traction across the domains of augmented and virtual reality (AR/VR), smart home automation, asset tracking, indoor navigation, and indoor drone technologies [1]. Many emerging advancements in these domains require accurate, lightweight, and real-time location tracking on resource-limited edge and mobile devices [2]. More broadly, the indoor positioning and navigation market that was valued at USD 10.9 billion in 2023 is expected to reach USD 29.8 billion by 2028, underscoring the critical role of these systems [3]. One of the most widely used methods for indoor localization is based on Wi-Fi received signal strength (RSS) fingerprinting, which utilizes the strength of signals from Wi-Fi access points (APs) to determine the position of mobile devices indoors [4]. This approach benefits from the ubiquity of Wi-Fi infrastructure and its compatibility with modern smart mobile devices, making it a favorable solution for real-world deployments.

Wi-Fi RSS fingerprinting operates by collecting RSS values from available APs at specific locations, forming a "fingerprint" unique to that position. As mobile devices move across different reference points (RPs) within a building, these RSS values fluctuate due to environmental factors and changes in position [4], [5]. Machine learning (ML) models are trained on these patterns, allowing them to map RSS fingerprints to precise locations. However, device heterogeneity—the differences in hardware, software, and firmware across mobile devices—introduces additional variability in the RSS measurements, complicating the localization process [5]. Furthermore, environmental factors such as multipath fading, shadowing, and human interference exacerbate this variability, making it increasingly challenging for ML models to maintain accurate predictions [6], [28], [29]. A traditional solution to address these challenges involves extensive data collection across a wide variety of devices and environmental conditions to ensure that the ML model learns to handle this variability [2], [4], [26], [27]. However, such a comprehensive data collection process is resource-intensive and impractical in real-world scenarios. A more efficient approach is federated learning (FL), which allows user-carried mobile devices (referred to as clients) to train local models (LMs) using their collected RSS fingerprints and collaboratively update a global model (GM) at a central server [7]. This approach helps FL systems adapt to heterogeneous environments and devices while reducing manual data collection efforts and maintaining user privacy by ensuring that sensitive data remains local to the client devices.

Despite these advantages, FL-based indoor localization systems are vulnerable to data poisoning attacks, where malicious clients intentionally inject poisoned or manipulated data into the GM's learning process [8]. Such attacks severely degrade the performance of the GM, as the erroneous LM updates from compromised devices are aggregated into the GM, leading to inaccurate localization predictions. To illustrate the impact of data poisoning attacks on FL-based indoor localization, we conducted experiments on two state-of-the-art FL-based indoor localization frameworks, FEDHIL [9] and FEDLOC [10], both of which use a three-layer deep neural network (DNN) as their GM. However, the aggregation strategies they employ differ: FEDLOC uses the traditional federated averaging (FEDAvg) method [11] to aggregate the LM updates, while FEDHIL utilizes a domain-specific selective weight aggregation technique that averages only specific weight tensors to mitigate bias from individual clients (discussed further in Section II). We subjected both frameworks to two types of data poisoning attacks: label flipping, where the labels are randomly altered, and backdoor attacks, implemented using the popular Fast Gradient Sign Method (FGSM) [12] (discussed further in Section III).

In Fig. 1, we present the performance degradation caused by these attacks, showing worst case errors (upper whisker), best case errors (lower whisker), and mean errors (center bar). Under label flipping, FEDLOC exhibits a 3.5× increase in mean localization errors, while backdoor attacks cause a staggering 6.5× rise in mean errors. FEDHIL, due to its selective aggregation strategy, shows relatively better resilience to backdoor attacks (3.25× increase in mean errors) but slightly worse performance for label flipping attacks (3.9× increase in mean errors). To some extent, the results highlight the role of aggregation strategies in defending against data poisoning attacks in FL systems, but in general both frameworks suffer from high localization errors. Moreover, many existing solutions for mitigating data poisoning rely on bulky, resource-intensive models that are unsuitable for real-time deployment on resource-limited edge and mobile devices (see Section II), where lightweight models and low inference latency are critical requirements.

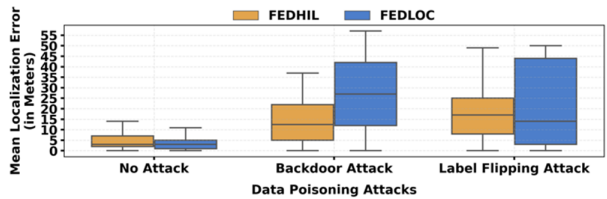

**Fig. 1.** Indoor localization error comparison for FEDLOC [10] and FEDHIL [9] under label flipping and backdoor poisoning attacks.

To address these multi-faceted challenges, we propose *SAFELOC*, a novel FL-based indoor localization framework that mitigates uncertainties due to client device heterogeneity and ML data poisoning attacks, while ensuring lightweight deployment on mobile devices. Our novel contributions are:

- We propose a new fused neural network architecture that performs poison detection, poison de-noising, and localization within the same model, and is optimized for efficient deployment on resource-limited devices.
- We propose a novel saliency map-based aggregation strategy that adjusts the influence of LMs based on the severity of detected poisoning, ensuring robust protection against ML data poisoning attacks.
- We perform extensive analysis on real-world data to evaluate *SAFELOC* under different data poisoning attacks and heterogeneous mobile devices, across various buildings, and contrast our performance against multiple state-of-the-art approaches.

## II. RELATED WORK

FL was first introduced by Google in 2016 to address privacy concerns and reduce the communication costs associated with training ML models across decentralized devices [13]. By allowing devices to collaboratively learn from local data without sharing it, FL enhances privacy and reduces bandwidth requirements. This paradigm shift has gained traction in various fields, particularly in indoor localization systems, where FL has become popular for its ability to reduce the data collection efforts, preserve privacy, and adapt to variations in RSS due to device heterogeneity and environmental conditions [14].

Traditional non-FL-based solutions like DNNLOC [15], VITAL [16], TIPS [17], STELLAR [6], and ANVIL [18] have shown potential for indoor localization, particularly in addressing variations due to device heterogeneity. However, these solutions often require large-scale data collection to train the ML models effectively. To alleviate this burden, frameworks such as SANGRIA [19], DATALOC [20], and WiDeep [21] employ data augmentation techniques to improve model performance in heterogeneous environments. Despite these efforts, these models are static and cannot adapt in real-time to environmental changes, limiting their effectiveness in practical indoor settings. In contrast, FL-based approaches can better adapt to environmental changes and can also minimize data collection overheads.

One of the early FL-based solutions in indoor localization was KRUM [22], which employed a simple Multi-Layer Perceptron (MLP) as the GM and used Euclidean distance-based filtering to select the LM update that deviated the least from the majority. While KRUM [22] reduces data collection overheads, it fails to incorporate collaborative learning from all clients, limiting its resilience to device heterogeneity effects. FEDLOC [10] improved upon KRUM by adopting a deep neural network (DNN) as the GM and utilizing FEDAvg for LM weight aggregation. While FEDAvg allows for collaborative learning and improves heterogeneity resilience, it introduces biases from noisy LMs, which can degrade GM performance. To address this, FEDHIL [9] introduced a DNN-based GM with a domain-specific aggregation strategy that selectively incorporated relevant weight tensors from LMs. This approach improves the robustness of the GM by reducing the impact of noisy or erroneous updates. However, despite these advancements, all of these methods remain vulnerable to data poisoning attacks, where malicious clients alter their local RSS data to produce erroneous LM updates that, when aggregated, corrupt the GM.

While no prior work has yet proposed a solution for addressing data poisoning attacks for indoor localization solutions, there have been some efforts for data poisoning attack mitigation in general FL systems. FEDCC [23] addresses data poisoning in FL systems by employing clustering techniques to group LMs based on gradient similarity, allowing it to detect and exclude poisoned updates from the GM aggregation. However, this approach may inadvertently filter out legitimate updates, particularly in heterogeneous environments. To further refine defense mechanisms, FEDLS [24] employs autoencoder-based latent space representations to detect anomalous LM updates. While this method enhances the detection of subtle poisoning attacks, it is resource-intensive, limiting its deployment on mobile devices. Similarly, ONLAD [25], another anomaly detection framework, uses a semi-supervised autoencoder to detect poisoning on mobile devices. However, both FEDLS and ONLAD face challenges due to their high resource demands, as they employ two separate ML models for poison detection and localization. Additionally, both frameworks rely on FEDAvg for aggregation, which still introduces biases from noisy LMs into the GM.

From the analysis of prior works, it is evident that existing solutions either fail to address the challenges of device heterogeneity and data poisoning, or introduce significant computational overhead, particularly in the presence of data poisoning attacks. To the best of our knowledge, our proposed FL-based framework (*SAFELOC*) is the first to address data poisoning attacks in the indoor localization domain. The framework also co-addresses uncertainties arising from heterogeneous client mobile devices. A fused neural network architecture is introduced to perform poison detection, poison de-noising, and localization within a single model, significantly reducing computational overhead without compromising accuracy. Additionally, a novel saliency map-based aggregation technique helps invert the effects of poisoned weight tensors, enhancing localization performance further, across diverse real-world scenarios.

## III. Data poisoning attacks in FL indoor localization

While FL systems offer significant privacy and collaboration benefits, they are susceptible to various forms of data poisoning attacks. These attacks are particularly harmful in FL-based indoor localization frameworks, as they allow malicious clients to deliberately manipulate their LMs and degrade the overall performance of the GM, which leads to inaccurate location predictions.

In a typical FL-based indoor localization framework, as depicted in Fig. 2, a central server maintains a GM that is distributed to several clients (mobile devices). Each client collects local RSS data from nearby APs and uses it to predict its location. The predicted label and local RSS data are then used to re-train the GM copy on the device, generating an LM. These LMs are uploaded back to the server, where an aggregation phase updates the GM based on the received LMs. This iterative process continues to improve localization performance as more local data is incorporated into the GM. However, among all the clients, there may be malicious clients who submit poisoned LM updates to the server. These poisoned updates, if integrated into the GM, can degrade the entire system's accuracy.

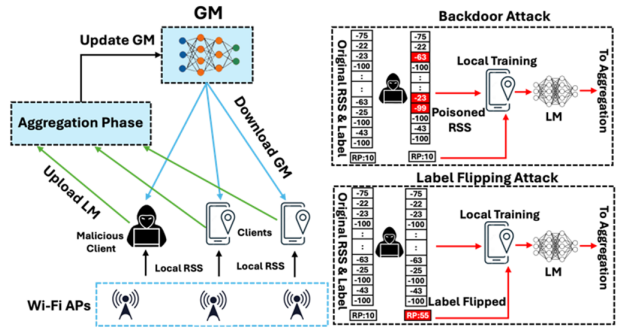

**Fig. 2.** Working of an FL-based indoor localization framework with malicious clients (with backdoor and label-flipping attacks).

Data poisoning attacks in FL can generally be classified into two types: backdoor attacks and label-flipping attacks. While both target the integrity of the LMs before they are uploaded to the server, they differ in their execution and overall impact:

- **Backdoor Attacks:** Here, a malicious client deliberately poisons its local RSS fingerprints to train a poisoned LM. This LM, when uploaded to the server, influences the GM to behave incorrectly for specific inputs.
- **Label-Flipping Attacks:** Here, instead of modifying RSS fingerprints, the attacker retains correct RSS data but intentionally flips labels associated with the location data. As a result, LM updates are trained to associate valid RSS fingerprints with the wrong locations.

### A. Data Poisoning Attack Methods

We consider four different methods to generate backdoor attacks within FL-based indoor localization systems: Clean Label Backdoor (CLB), Fast Gradient Sign Method (FGSM), Projected Gradient Descent (PGD), and Momentum Iterative Method (MIM). Additionally, we also consider label-flipping attacks. These five methods aim to degrade localization accuracy by altering RSS fingerprints or labels during local training. These techniques are described below:

- **Clean Label Backdoor (CLB)**: The CLB attack inserts perturbations into the LM by modifying the RSS fingerprints while leaving the label unchanged. The attack calculates a mask value along with the perturbation strength to generate poisoned samples. The CLB attack can be described as follows:

$$X_{CLB} = X + \epsilon * \delta(\nabla J(X,Y)) \quad (1)$$

where, $X$ is the local RSS fingerprint, $\epsilon$ is the perturbation magnitude, and $\delta$ is the mask value that is computed using the gradients of the GM's loss function $\nabla J(X,Y)$. This generates a new RSS fingerprint $X_{CLB}$, designed to poison the GM through aggregation.

- **Fast Gradient Sign Method (FGSM)**: The FGSM attack is a one-step, non-iterative attack method that calculates perturbations using the gradient of the GM's loss function with respect to the local RSS data. The FGSM attack is implemented using the equation below:

$$X_{FGSM} = X + \epsilon * sign(\nabla J(X,Y)) \quad (2)$$

where $\epsilon$ is the perturbation magnitude, and $\nabla J(X,Y)$ represents the gradient of the GM's loss function with respect to the local RSS fingerprint $X$. This perturbation generates a new RSS fingerprint $X_{FGSM}$, designed to poison the GM through aggregation.

- **Projected Gradient Descent (PGD)**: The PGD attack is an iterative method that generates perturbations by applying small updates to the local RSS based on the GM's gradient. It is an iterative version of FGSM and ensures that the perturbation remains within a predefined bound. The PGD attack is given by:

$$X_{PGD} = X + Proj_{X,\epsilon}\{\epsilon * \frac{\nabla J(X,Y)}{L|\nabla J(X,Y)|_2}\} \quad (3)$$

where $Proj_{X,\epsilon}$ ensures that the induced perturbation remains within a certain bound ($X,\epsilon$), $\nabla J(\theta,X,Y)$ represents the gradient of the GM's loss function and $L|\nabla J(\theta,X,Y)|_2$ represents the squared L2 norm (ridge regularization) of the gradients of the GM.

- **Momentum Iterative Method (MIM)**: The MIM attack builds on PGD by incorporating momentum into the gradient updates, making the attack more effective by maintaining direction across iterations. The MIM attack can be described as follows:

$$X_{MIM} = \alpha * X + Proj_{X,\epsilon}\{\epsilon * \frac{\nabla J(X,Y)}{L|\nabla J(X,Y)|_2}\} \quad (4)$$

where, $\alpha$ represents the momentum term. This method often leads to very potent data poisoning samples.

- **Label-Flipping**: The label-flipping attack leaves the local RSS data $X$ unmodified but flips its corresponding label $y$ to an incorrect class $y_{FLIP}$, which leads the LM to associate valid RSS data with incorrect locations. The label-flipping attack can be described as follows:

$$y_{FLIP} = FLIP(y) \quad (5)$$

where $FLIP(y)$ denotes the operation that switches the original label $y$ to an incorrect class $y_{FLIP}$.

## IV. SAFELOC Framework

The *SAFELOC* framework consists of two primary components: a fused neural network as the GM and a saliency map-based aggregation strategy to mitigate the effects of data poisoning attacks. The framework starts with training the fused neural network on a centralized server using a subset of RSS fingerprints from the building floorplan. This fused neural network integrates an autoencoder for poison detection and poison de-noising, alongside a classification layer for location prediction. By sharing layers between the autoencoder and classification layers, computational overhead is reduced, making the model suitable for resource-limited mobile device deployment. During training, the autoencoder encodes the RSS fingerprints into a low-dimensional latent space and reconstructs them using the de-noising decoder. The reconstruction error (RCE) is calculated as the difference between the original and reconstructed RSS data. This error is used to establish a threshold ($\tau$) for detecting poisoned data during client-side training. Simultaneously, the classification layer learns to map RSS fingerprints to corresponding location labels. After training, the GM is distributed to clients, who use it to evaluate their local RSS data.

The modified autoencoder first checks for backdoor poisoning by comparing the RCE of the local RSS fingerprints to $\tau$ set during initial training. If the RCE $<= \tau$, the data is considered clean, and the classification layer generates location predictions. The client then retrains its LM on this data and uploads it to the server for aggregation, to globally improve indoor localization accuracy for all clients. If the RCE $> \tau$, the RSS data is flagged as potentially poisoned via a backdoor attack. The reconstructed data, obtained by the de-noising decoder, is passed through the classification layer to generate location coordinates. In this case, the de-noising decoder removes (by de-noising) the backdoor poisoned perturbations, and the LM is uploaded to the server.

In the case of label-flipping attacks, the attacker flips the predicted location coordinates before updating the LM, causing the wrong class to be updated even though the RSS fingerprint remains clean. When this poisoned LM is sent to the server, it can cause LM weight tensors to be significantly different from that of the GM. To counter this, a saliency map-based aggregation strategy is designed to compare the LM's weight tensors with those of the GM. A saliency map matrix is created where similar tensors are assigned high saliency values, and highly deviated tensors are assigned low values. The saliency map is then used to adjust the LM weight tensors before performing aggregation with the GM.

In the following subsections, we describe the working of the fused neural network and saliency map-based aggregation strategy used in the *SAFELOC* framework.

*A. Fused Neural Network Architecture*

The fused neural network in *SAFELOC* integrates an autoencoder for poison detection and de-noising alongside a classification layer for location prediction, as seen in Fig. 3. The autoencoder is composed of an encoder and a de-noising decoder. The encoder compresses the RSS fingerprint data into a progressively smaller latent space using fully connected layers, reducing the number of neurons at each layer until the bottleneck layer. This bottleneck represents a low-dimensional latent space that captures the key features of the input data. The latent space is then bifurcated into two outputs: one directed to the de-noising decoder for reconstruction and the other to the classification layer for location prediction.

The de-noising decoder uses fully connected layers with an increasing number of neurons, structured similarly to the encoder's architecture but in reverse, to reconstruct the original RSS fingerprint. To improve precision of reconstruction, we freeze the gradients from the encoder and propagate them to their corresponding layers in the decoder. This approach helps maintain the learned patterns during the encoding process while ensuring the decoder properly removes any perturbations, particularly from backdoor poisoning attacks. The RCE is calculated using the Mean Squared Error (MSE) between the input RSS fingerprint and the reconstructed RSS fingerprint. If RCE $<= \tau$, the latent space representation is forwarded to the classification layer for location prediction. If the RCE $> \tau$, the reconstructed fingerprint undergoes another encoding step (re-supplied to the encoder as input) to generate a new latent space, which is then passed to the classification layer for location predictions. This approach mitigates backdoor poisoning by removing perturbations during the reconstruction phase. When label-flipping attacks occur, the input RSS data remains clean, but the label is manipulated before the LM update. This situation is addressed in the server through the saliency map-based aggregation method, which adjusts LM weight tensors to safeguard the GM from poisoning, as discussed next.

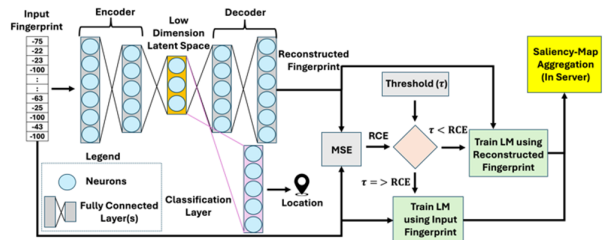

**Fig. 3.** Fused neural network architecture used in *SAFELOC*.

*B. Saliency-Map Based Aggregation*

The saliency map-based aggregation strategy is used to mitigate the effects of poisoned LMs by adjusting their weight tensors before aggregation. Once the server receives the LMs, it first computes a deviation matrix ($\Delta W_i$) to quantify the difference between the weight tensors of the LM and the GM, which is calculated as:

$$\Delta W_i = |W_{LM,i} - W_{GM,i}| \quad (6)$$

where *i* represents the index of the individual weight tensors, $W_{LM,i}$ represents the weight tensors of the LM, and $W_{GM,i}$ represents the weight tensors of the GM. To counter the influence of these deviations, we calculate a saliency matrix ($S_i$) for each weight tensor using the inverse deviation method, as shown in the equation:

$$S_i = \frac{1}{1 + \Delta W_i} \quad (7)$$

The $S_i$ adjusts the magnitude of each weight tensor in $W_{LM,i}$ before aggregation. It assigns higher saliency values to weight tensors that closely match $W_{GM,i}$ (i.e., small deviation) and lower saliency values to weight tensors that show significant deviation. This ensures that large deviations are given less weight during aggregation. Once the $S_i$ is computed, it is used to adjust the $W_{LM,i}$ before aggregation. The adjusted LM weight tensors ($W_{Adj}$) are obtained by multiplying the $W_{LM,i}$ with the $S_i$ values, as shown below:

$$W_{Adj} = S_i * W_{LM,i} \quad (8)$$

This results in an adjusted LM that accounts for deviations while mitigating the potential influence of poisoned weights.

The $W_{Adj}$ is then aggregated with the GM using:

$$W'_{GM} = W_{GM} + W_{Adj} \qquad (9)$$

where $W'_{GM}$ represents the updated GM. This process reduces the impact of poison updates on the GM, ensuring that the GM remains robust against poisoned LM updates.

## V. EXPERIMENTAL RESULTS

### A. Experimental Setup

In our experiments, we evaluate the performance of the *SAFELOC* framework across *1)* real-world RSS fingerprint data from different buildings, *2)* various data poisoning attack scenarios, and *3)* contrast performance with state-of-the-art frameworks, including FEDHIL [9], FEDLOC [10], FEDCC [23], FEDLS [24], and ONLAD [25].

We collected RSS fingerprint data from six different client mobile devices (Samsung Galaxy S7, OnePlus 3, Motorola Z2, LG V20, BLU Vivo 8, HTC U11). Data was collected along paths from five different buildings, with each path having different numbers of reference points (RPs) and different numbers of visible APs. Building 1 comprises 60 RPs with up to 203 visible APs, Building 2 consists of 48 RPs with 201 visible APs, Building 3 includes 70 RPs with 187 visible APs, Building 4 has 80 RPs with 135 visible APs, and Building 5 features 90 RPs with 78 visible APs. The granularity for all RPs is set at 1 meter, which we consider sufficient for indoor localization. We standardized the RSS values between 0 dBm (strongest signal) and -100 dBm (weakest signal). Training was done on data collected from one device (Motorola Z2), and testing on the remaining five devices. Training involved collecting five fingerprints per RP per building, while testing used one fingerprint per RP per building. This method strikes a balance between gathering sufficient training data and the practicality of data collection across real-world environments.

The *SAFELOC* framework was configured with specific hyperparameters: the encoder in the fused neural network consists of three fully connected layers with 128, 89, and 62 neurons, while the decoder has two fully connected layers with 89 and 128 neurons. ReLU activation was applied to all layers, and the autoencoder was trained using the MSE loss function. The classification layer consists of neurons corresponding to the number of RPs for each building. Sparse categorical cross-entropy was set as the classification layer's loss function. Both the autoencoder and the classification layer were trained using the Adam optimizer with a learning rate of 0.001 for 700 epochs. For client-side training, a reduced learning rate of 0.0001 and 5 epochs were used to facilitate lightweight local training.

In the following subsections, we present results of experiments to determine the optimal reconstruction threshold ($\tau$), examine *SAFELOC*'s performance under various data poisoning attacks and attack intensities, and compare *SAFELOC*'s localization accuracy with current state-of-the-art frameworks. We also present data on *SAFELOC*'s model inference latency and parameter footprint, and an analysis of the framework's scalability.

### B. Determination of Reconstruction Threshold ($\tau$)

As a first step, we analyzed the effect of varying the reconstruction threshold ($\tau$) on localization performance in *SAFELOC*. $\tau$ plays a crucial role in determining whether local RSS fingerprints collected by clients contain backdoor poisoning, based on the reconstruction error (RCE). We vary $\tau$ from 0 to 0.5 to find the optimal threshold that minimizes the mean localization error across the five buildings, calculated by averaging results from all six devices and RPs. This threshold allows variance for device heterogeneity and identify the most effective $\tau$ value for poison detection. The HTC U11 device is used to introduce data poisoning attacks (CLB, FGSM, PGD, MIM, and label flipping), with perturbation strength ($\epsilon$) varying from 0 to 1 in steps of 0.01 up to 0.1, then in steps of 0.1. Fig. 4 shows that as $\tau$ increases from 0.05 (5% tolerance) to 0.5 (50% tolerance), localization errors vary. The lowest mean error is achieved at $\tau = 0.1$, particularly for Building 4, where error variance is minimal. Localization errors remain stable between $\tau = 0.15$ and 0.25, but beyond $\tau = 0.3$, there is a significant increase in errors, peaking between $\tau = 0.45$ and 0.5. This indicates that higher $\tau$ values allow more poisoned RSS data to influence the GM, leading to larger errors. Based on these observations, we determine that $\tau = 0.1$, allowing a 10% variance, is optimal for minimizing localization errors while effectively detecting poisoned data. This $\tau$ will be used in subsequent evaluations under different poisoning scenarios.

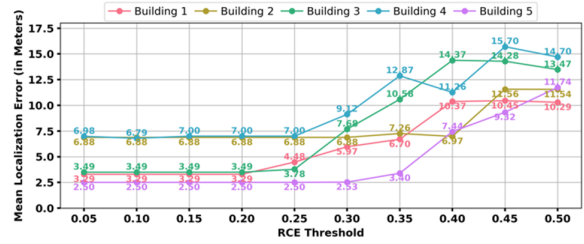

**Fig. 4:** Impact of varying the reconstruction threshold ($\tau$) on mean localization error across five building floorplans.

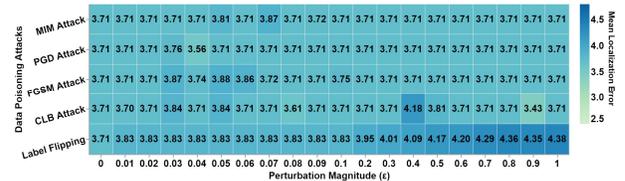

**Fig. 5.** Mean localization error under various data poisoning attacks and perturbation magnitudes ($\epsilon$).

### C. Effects of Data Poisoning Attacks in SAFELOC

Next, we investigate the impact of various data poisoning attacks on the *SAFELOC* framework, focusing on both low ($\epsilon$ = 0.01 to 0.09, indicating 1% to 9%) and high ($\epsilon$ = 0.1 to 1.0, indicating 10% to 100%) perturbation strengths, as depicted in Fig 5. Each row in the heatmap represents a different data poisoning attack, and each column represents a different $\epsilon$ value. The HTC U11 device is again used to introduce data poisoning attacks. The corresponding mean localization error across all devices, buildings, and RPs is recorded in each cell in the figure. It can be observed that up to $\epsilon < 0.1$, *SAFELOC* maintains a stable mean localization error across all attacks, even as perturbation magnitude increases. This stability indicates that the autoencoder and saliency map-based aggregation can effectively handle small perturbations. As we shift to $\epsilon > 0.1$, the mean errors remain stable for backdoor poisoning attacks, but for label-flipping attacks, we observe a slight increase in mean errors beginning at $\epsilon = 0.2$, reaching up to 4.38 meters at $\epsilon = 1.0$. The saliency map method, although highly effective, may assign relatively higher saliency to these poisoned weight tensors, leading to a slight error increase. Despite the small rise in mean errors at higher $\epsilon$ values the *SAFELOC* consistently outperforms state-of-the-art frameworks, as discussed next.

*D. Comparision with State-of-the-art*

We compared the performance of the *SAFELOC* framework against several state-of-the-art indoor localization frameworks. The box and whisker plot in Fig. 6 shows worst-case errors (upper whisker), best-case errors (lower whisker), and mean errors (center bar) across various poisoning scenarios. The results are aggregated across all buildings.

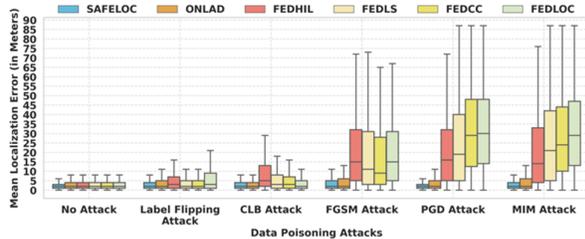

**Fig. 6.** Comparison of *SAFELOC* against state-of-the-art.

We observe that *SAFELOC* consistently achieves lower mean and worst-case localization errors compared to all the other frameworks. ONLAD ranks second due to its distinct poison detection and localization architecture, but its use of FEDAvg allows noisy weight tensors to degrade the GM, particularly for label flipping, FGSM, and MIM attacks. FEDLS performs well under label flipping attacks but suffers in backdoor attacks as its autoencoder latent space is vulnerable to stronger perturbations. FEDCC, while resilient to label flipping, is weak against backdoor attacks, as it inadvertently filters legitimate LMs under severe perturbations. FEDHIL's selective weight aggregation aggregates large tensor changes caused by attacks, leading to higher errors, particularly in backdoor scenarios. Lastly, FEDLOC exhibits the highest localization errors across all attack types due to its lack of defense mechanisms. *SAFELOC* achieves 1.2× to 2.11× lower mean errors and 1.16× to 2.33× lower worst-case errors for label flipping, and 1.33× to 5.9× lower mean errors and 1.5× to 7.8× lower worst-case errors for backdoor attacks (CLB, FGSM, PGD, MIM) compared to other frameworks.

Table I shows the implementation overheads of all compared frameworks. *SAFELOC* has the least model parameters and the lowest model inference latency, offering 1.04× to 2.1× faster inference than the state-of-the-art frameworks.

TABLE I : MODEL LATENCY AND PARAMETERS COMPARISON

| Framework | Model Inference Latency | Total Parameters |
|---|---|---|
| SAFELOC | 64 ms | 41,094 |
| ONLAD | 87 ms | 130,185 |
| FEDHIL | 84 ms | 97,341 |
| FEDCC | 67 ms | 42,993 |
| FEDLS | 103 ms | 282,676 |
| FEDLOC | 135 ms | 137,801 |

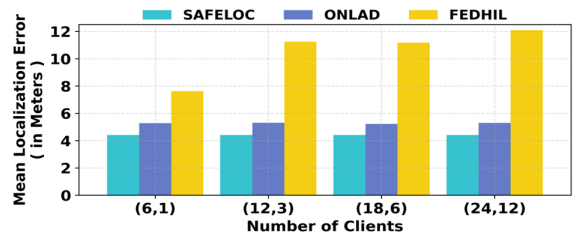

**Fig. 7.** Mean localization error with increasing number of (total, poisoned) clients ratio.

*E. Scalability Analysis with Increasing number of Clients*

Lastly, we evaluate *SAFELOC's* scalability against the two best frameworks from prior works (ONLAD, FEDHIL) by increasing clients from 6 to 24, with poisoned clients rising from 1 to 12. Fig 7 shows FEDHIL exhibits a steady rise in mean localization errors as more clients are introduced, particularly with an increase in poisoned clients, while ONLAD and *SAFELOC* maintain stable performance. *SAFELOC* consistently shows the lowest mean localization errors, owing to its saliency map aggregation that mitigates the impact of poisoned clients, even as their number grows.

## VI. CONCLUSIONS

In this paper, we described *SAFELOC*, a new framework that effectively mitigates the impact of various data poisoning attacks in heterogeneous federated learning based indoor localization environments. The novel fused neural network and saliency map-based aggregation in *SAFELOC* allow it to achieve up to 5.9× better mean localization error and 2.1× faster model inference latency compared to state-of-the-art frameworks, making it efficient for real-world deployment.